\documentclass[preprint,12pt]{elsarticle}

\newcommand{\review}[1]{\textcolor{black}{#1}}

\usepackage{graphicx}

\usepackage{amsmath, amsthm, amssymb}

\usepackage{lineno}
\usepackage{xspace}
\usepackage{color}

\newcommand{\HC}{Hill Climbing\xspace}
\newcommand{\TS}{Tabu Search\xspace}
\newcommand{\GA}{Genetic Algorithms\xspace}

\journal{Complexity}

\begin{document}

\begin{frontmatter}


\title{Learning the structure of Bayesian Networks: A quantitative assessment of the effect of different algorithmic schemes}



\author[unimib]{Stefano Beretta}
\ead{stefano.beretta@disco.unimib.it}
\author[novaims]{Mauro Castelli}
\ead{mcastelli@novaims.unl.pt}
\author[coimbra]{Ivo Gon\c{c}alves}
\ead{icpg@dei.uc.pt}
\author[novaims]{Roberto Henriques}
\ead{roberto@novaims.unl.pt}
\author[stanford]{Daniele Ramazzotti\corref{cor1}}
\ead{daniele.ramazzotti@stanford.edu}
\cortext[cor1]{Corresponding author}

\address[unimib]{DISCo, Universit\'a degli Studi di Milano-Bicocca. 20126 Milano, Italy}
\address[novaims]{NOVA Information Management School (NOVA IMS), Universidade Nova de Lisboa, Campus de Campolide, 1070-312 Lisboa, Portugal}
\address[coimbra]{INESC Coimbra, DEEC, University of Coimbra, Portugal}
\address[stanford]{Department of Pathology, Stanford University. California, USA}

\begin{abstract}
One of the most challenging tasks when adopting Bayesian Networks
(BNs) is the one of learning their structure from data.
This task is complicated by the huge search space of possible
solutions, and by the fact that the problem is $NP$-hard. Hence, \review{full enumeration of all the possible solutions is not always feasible and approximations are often required}.
However, to the best of our knowledge, a quantitative analysis of the performance and
characteristics of the different heuristics to solve this problem
has never been done before. 

For this reason, in this work, we provide a detailed \review{comparison} of \review{many} different state-of-the-arts methods for structural learning on
simulated data considering both BNs with discrete and continuous variables, and with different rates of noise in the data.
In particular, we investigate the \review{performance} of different
widespread scores \review{and algorithmic approaches} proposed for the inference and the statistical pitfalls within them.
\end{abstract}


\end{frontmatter}


\section{Introduction}
\label{sec:intro}

Bayesian Networks (BNs) have been applied to several different fields,
ranging from the water resources management~\cite{leu2016leak} to the
discovery of gene regulatory
networks~\cite{dondelinger2013non,young2014fast}.
The task of learning a BN can be divided into two subtasks:
(1) structural learning, i.e., identification of the topology of
the BN, and (2) parametric learning, i.e., estimation of the
numerical parameters (conditional probabilities) for a given network
topology.
In particular, the most challenging task of the two is the one of
learning the structure of a BN.
Different methods have been proposed to face this problem, and
they can be classified into two
categories~\cite{buntine1996guide,daly2011,koller2009probabilistic}:
(1) methods based on detecting conditional independences, also known
as constraint-based methods, and (2) score+search methods, also known
as score-based approaches.
\review{It must be noticed that hybrid methods have also been proposed
in~\cite{saptawati2005hybrid} but, for the sake of clarity, here we limit our
discussion to the two mainstream approaches to tackle the task.}

As discussed in~\cite{Larranaga2013109}, the input of the former
algorithms is a set of conditional independence relations between
subsets of variables, which are used to build a BN that represents
a large percentage (and, whenever possible, all) of these
relations~\cite{spirtes2000causation}.
However, the number of conditional independence tests that such
methods should perform is exponential and, thus, approximation
techniques are required. 

Although constraint-based learning is an interesting approach, as it
is close to the semantic of BNs, most of the developed structure
learning algorithms fall into the score-based method category,
given the possibility of formulating such a task in terms of an
optimization problem.
As the name implies, these methods have two major components: (1) a
scoring metric that measures the quality of every candidate BN with
respect to a dataset, and (2) a search procedure to intelligently
move through the space of possible networks, as this space is
enormous.
More in detail, as shown
in~\cite{chickering1996learning,chickering2004large}, searching
this space for the optimal structure is an $NP$-hard problem,
even when the maximum number of parents for each node is constrained.

Hence, regardless of the strategy to learn the structure, one wishes
to pursue, greedy local search techniques and heuristic search
approaches need to be adopted to tackle the problem of inferring
the structure of BNs.
However, to the best of our knowledge, a quantitative analysis of
the performance of these different techniques has never been done.
For this reason, here we provide a detailed assessment of the
performance of different state-of-the-arts methods for structural
learning on simulated data, considering both BNs with discrete and
continuous variables, and with different rates of noise in the data.
More precisely, we investigate the characteristics of different
scores proposed for the inference, and the statistical pitfalls
within both the constraint-based and score-based techniques.
Furthermore, we study the impact of various heuristic search
approaches, namely \HC, \TS, and \GA. 

\review{We notice that, although we here aim at covering some of the main ideas in the area of structure learning of BNs, several interesting topics are beyond the scope of this work~\cite{drton2017structure}. In particular, we refer to a more general formulation of the problem~\cite{koller2009probabilistic}, and we do not consider, for example, contexts where it is possible to exploit prior knowledge in order to make the tasks computationally affordable~\cite{stefanini2014chain}. At the same time, in this work we do not investigate in details performance related to causal interpretations of the BNs~\cite{heinze2018causal}.}

This work is structured as follows.
In the next two Sections, we provide a background on both Bayesian
Networks and Heuristic Search Techniques (see 
Section~\ref{sec:background} and Section~\ref{sec:exp}).
In Section~\ref{sec:results} we describe the results of our study
and in Section~\ref{sec:conclusions} we conclude the paper. 

\section{Bayesian Networks}
\label{sec:background}

BNs are graphical models representing the \emph{joint distribution}
over a set of $n$ random variables through a \emph{direct acyclic
graph} (DAG) $G=(V,E)$, where the $n$ nodes in $V$ represent the
variables, and the arcs $E$ encode any statistical dependence
among them.
\review{Similarly, the lack of arcs among variables subsumes statistical independence}.
In this DAG, the set of variables with an arc toward a given node
$X \in V$ is its set $\pi(X)$ of ``parents''.
\review{Formally, a Bayesian Network \cite{koller2009probabilistic} is defined as a pair $\langle G,\theta \rangle$ over the variables $V$, with arcs $E\subseteq V\times V$ and real-valued parameters $\theta$.} 
When the structure of a BN is known, it is possible to compute the
joint distribution of all the variables as the product of the
conditional distributions on each variable given its parents. 

\[
p(X_1,...,X_n) = \prod_{X_i=1}^n p(X_i|\pi(X_i)) \qquad p(X_i|\pi(X_i))=\theta_{X_i|\pi(X_i)}\, ,
\]

\review{where $\theta_{X_i|\pi(X_i)}$ is a {\em probability density function}.} 

However, even if we consider the simplest case of binary variables,
the number of parameters in each conditional probability table is
still exponential in size.
For example, in the case of binary variables, the total number of
parameters needed to compute the full joint distribution is of size
$\sum_{X \in V} 2^{|\pi(X)|}$, where $|\pi(X)|$ is the cardinality of
the parent set of each node.
\review{Notice that, if a node does not have parents the total number of
parameters to be computed is $1$, which corresponds to its
marginal probability.}

\review{Moreover, the usage of the symmetrical notion of
conditional dependence poses further limitations to the task of
learning the structure of a BN: two arcs $A \rightarrow B$ and 
$B \rightarrow A$ in a network can in fact equivalently denote dependence between 
variables $A$ and $B$; this leads to the fact that two DAGs 
having a different structure can sometimes model an identical set of independence 
and conditional independence relations (\emph{I-equivalence}). This yields to the
concept of \emph{Markov equivalence class} as a \emph{partially 
directed acyclic graph} where the arcs that can take either
orientation are left undirected \cite{koller2009probabilistic}. In this 
case, all the structures within the Markov equivalence class are 
equivalently ``good'' in representing the data, unless a causal interpretation 
of the BN is given \cite{judea1991equivalence}.}

In the literature, there are two families of methods to learn
the structure of a BN from data.
The idea behind the first group of methods is the one of
\emph{learning the conditional independence relations} of the BN
from which, in turn, the network is learned.
These methods are often referred to as {\em constraint-based approaches}.
The second group of methods, the so-called  \emph{score-based
approaches}, formulates the task of structure learning as an
optimization problem, with scores aimed at {\em maximizing the
likelihood of the data} given the model.
However, both the approaches are known to lead to $NP$-hard formulations
and, because of this, heuristic methods need to be used to find
\review{near optimal solutions with high probability, in a reasonably
small number of iterations.} 

\subsection{Constraint-based approaches}
\label{sec:bn_learning_structural}

We now briefly describe the main idea behind this class of
approaches.
For a detailed discussion of this topic we refer
to~\cite{spirtes2000causation,tsamardinos2003algorithms}. 

This class of methods aims at building a graph structure to
reflect the dependence \review{and independence} relations in the data that match the
empirical distribution.
Notwithstanding, the number of conditional independence tests that
such algorithms would have to perform among any pair of nodes to
test all possible relations is exponential and, because of this,
the introduction of some approximations is required. 

We now provide some details on two constraint-based algorithms that
have been proven to be efficient and of widespread usage:
the {\em PC algorithm}~\cite{spirtes2000causation} and
the {\em Incremental Association Markov Blanket} (IAMB)~\cite{tsamardinos2003algorithms}.

\paragraph{The PC algorithm}
This algorithm~\cite{spirtes2000causation} starts with a fully
connected graph and, on the basis of pairwise independence tests,
it iteratively removes all the extraneous edges.
To avoid an exhaustive search of separating sets, the edges are ordered
to consider the correct ones early in the search.
Once a separating set is found, the search for that pair ends.
The PC algorithm orders the separating sets by increasing values
of size $l$, starting from $0$ (the empty set), until $l = n-2$
(where $n$ is the number of variables).
The algorithm stops when every variable has less than $l-1$
neighbors since it can be proven that all valid sets must have
already been chosen.
During the computation, the bigger the value of $l$ is, the larger
the number of separating sets must be considered.
However, \review{as $l$ gets big}, the number of nodes with
degree $l$ or higher must have dwindled considerably.
Thus, in practice, we only need to consider a small subset of all
the possible separating sets. 

\paragraph{Incremental Association Markov Blanket algorithm}
An other constraint-based learning algorithms uses the
Markov blankets~\cite{koller2009probabilistic} to restrict the
subset of variables to test for independence.
Thus, when this knowledge is available in advance, we do not need
to test a conditioning on all possible variables.
A widely used and efficient algorithm for Markov blanket discovery
is IAMB which, for each variable $X$, keeps track of a hypothesis
set ${\cal H}(X)$, \review{which is the set of nodes that may be parents of $X$.}
The goal is, for a given ${\cal H}(X)$, to obtain at the end of
the algorithm, a Markov blanket of $X$ equal to ${\cal B}(X)$.
IAMB consists of a forward and a backward phase.
During the forward phase, it adds all the possible variables into
${\cal H}(X)$ that could be in ${\cal B}(X)$, while in the backward
phase, it removes all the false positive variables from the
hypotheses set, leaving the true ${\cal B}(X)$.
The forward phase begins with an empty ${\cal H}(X)$ for each $X$.
Then, iteratively, variables with a strong association with $X$
(conditioned on all the variables in ${\cal H}(X)$) are added to
the hypotheses set.
This association can be measured by a variety of non-negative
functions, such as {\em mutual information\/}.
As ${\cal H}(X)$ grows large enough to include ${\cal B}(X)$, the
other variables in the network will have very little association
with $X$, conditioned on ${\cal H}(X)$.
At this point, the forward phase is complete.
The backward phase starts with ${\cal H}(X)$ that contains
${\cal B}(X)$ and false positives, which will have small conditional
associations, while true positives will associate strongly.
Using this test, the backward phase is able to iteratively remove
the false positives, until all but the true negatives are eliminated. 

\subsection{Score-based approaches}
\label{sec:bn_learning_score}
These approaches aim at maximizing the likelihood $\mathcal{L}$ of
a set of observed data $D$, which can be computed as the product
of the probability of each observation. Since we want to infer a model 
$G$ that best explains the observed data, we define the likelihood of 
observing the data given a specific model $G$ as: 

\[
\mathcal{LL}(G;D) = \prod_{d \in D} P(d|G)\, .
\]

Practically, however, for any arbitrary set of data, the most likely
graph is always the fully connected one, since adding an edge can
only increase the likelihood of the data, i.e.,~this approach overfits
the data.
To overcome this limitation, the likelihood score is almost always
combined with a {\em regularization term} that penalizes the
complexity of the model in favor of sparser
solutions~\cite{koller2009probabilistic}. 

As already mentioned, such an optimization problem leads to
intractability, due to the enormous search space of the valid solutions.
Because of this, the optimization task is often solved with
heuristic techniques.
Before moving on to describe the main heuristic methods employed
to face such complexity (see Section~\ref{sec:exp}), we now give
a short description of a particularly relevant and known score,
called {\em Bayesian Information Criterion}
(BIC)~\cite{schwarz1978estimating}, as an example of scoring
function adopted by several the score-based methods. 

\paragraph{The Bayesian Information Criterion}
BIC uses a score that consists of a log-likelihood term and a
regularization term that depend on a model $G$ and data $D$:
\begin{equation}
\label{eq:bic}
\textsc{bic}(G;D) = \mathcal{LL}(G;D) - \dfrac{\log m}{2} \text{dim}(G)
\end{equation}
where $D$ denotes the data, $m$ denotes the number of samples, and
$\text{dim}(G)$ denotes the number of parameters in the model.
\review{We recall that in this formulation, the BIC score should be maximized.}
Since, in general, $\text{dim}(\cdot)$ depends on the number of
parents of each node, it is a good metric for model complexity.
Moreover, each edge added to $G$ increases the complexity of the
model.
Thus, the regularization term based on $\text{dim}(\cdot)$ favors
graphs with fewer edges and, more specifically, fewer parents for
each node.
The term ${\log m}/{2}$ essentially weighs the regularization term.
The effect is that the higher the weight, the more sparsity will
be favored over ``explaining'' the data through maximum likelihood. 

Notice that the likelihood is implicitly weighted by the number of
data points since each point contributes to the score.
As the sample size increases, both the weight of the regularization
term and the ``weight'' of the likelihood increase.
However, the weight of the likelihood increases faster than that
of the regularization term.
This means that, with more data, the likelihood will contribute more
to the score, and we may trust our observations more and have less
need for regularization.
Statistically speaking, BIC is a {\em consistent score}~\cite{koller2009probabilistic}.
Consequently, $G$ contains the same independence relations as
those implied by the true structure.

\section{Heuristic Search Techniques}
\label{sec:exp}

We now describe some of the main state-of-the-art search strategies
that we took into account in this work.
In particular, as stated in Section~\ref{sec:intro}, we considered
the following search methods: \HC, \TS, and \GA.

\subsection{Hill Climbing}
\label{sec:hc}

\HC (HC) is one of the simplest iterative techniques that have
been proposed for solving optimization problems.
While HC consists of a simple and intuitive sequence of steps, it
is a good search scheme to be used as a baseline for comparing the
performance of more advanced optimization techniques.

\HC shares with other techniques (like Simulated
Annealing~\citep{hwang1988simulated} and \TS~\citep{glover1989tabu})
the concept of neighborhood.
Search methods based on this latter concept are iterative procedures
in which a neighborhood $N(i)$ is defined for each feasible solution
$i$, and the next solution $j$ is searched among the solutions in
$N(i)$.
Hence, the neighborhood is a function $N:S\rightarrow 2^S$ that assigns
to each solution in the search space $S$ a (non-empty) subset of $S$.

The sequence of steps of the \HC algorithm, for a \review{maximization}
problem w.r.t.~a given objective function $f$, are the following:
\begin{enumerate}
\item choose an initial solution $i$ in $S$;
\item find the best solution $j$ in $N(i)$ (i.e., the solution $j$ such
that $f(j)\geq f(k)$ for every $k$ in $N(i)$;
\item if $f(j) < f(i)$, then stop; else set $i=j$ and go to Step 2.
\end{enumerate}

As it is clear from the aforementioned algorithm, \HC returns
a solution that is a local \review{maximum} for the problem at hand.
This local \review{maximum} does not generally correspond to the global
minimum for the problem under exam, that is, \HC does not
guarantee to return the best possible solution for a given problem.
To counteract this limitation, more advanced neighborhood search methods
have been defined.
One of these methods is \TS, an optimization technique that uses
the concept of ``memory''.

\subsection{Tabu Search}
\label{sec:ts}
\noindent

\TS (TS) is a meta-heuristic that guides a local heuristic search procedure
to explore the solution space beyond local optimality.
One of the main components of this method is the use of an adaptive memory,
which creates a more flexible search behavior.
Memory-based strategies are therefore the main feature of TS approaches,
founded on a quest for ``integrating principles'', by which alternative forms
of memory are appropriately combined with effective strategies for exploiting
them. 

\emph{Tabus} are one of the distinctive elements of TS when compared to \HC or
other local search methods.
The main idea in considering \emph{tabus} is to prevent cycling when moving
away from local optima through non-improving moves.
When this situation occurs, something needs to be done to prevent the search
from tracing back its steps to where it came from.
This is achieved by declaring \emph{tabu} (disallowing) moves that reverse the
effect of recent moves.
For instance, let us consider a problem where solutions are binary strings of
a prefixed length, and the neighborhood of a solution $i$ consists of the
solutions that can be obtained from $i$ by flipping only one of its bits.
In this scenario, if a solution $j$ has been obtained from a solution $i$ by
changing one bit $b$, it is possible to declare a \emph{tabu} to avoid to flip
back the same bit $b$ of $j$ for a certain number of iterations (this number
is called the tabu tenure of the move).
\emph{Tabus} are also useful to help in moving the search away from previously
visited portions of the search space and, thus, perform more extensive
exploration.

As reported in~\citep{Gendreau:2010:HM:1941310}, tabus are stored in a
short-term memory of the search (the tabu list) and usually, only a fixed
limited quantity of information is recorded.
It is possible to store complete solutions, but this has a negative impact on
the computational time required to check whether a move is a tabu or not and,
moreover, it requires a vast amount of space.
The second option (which is the one commonly used) involves recording the last
few transformations performed to obtain the current solution and prohibiting
reverse transformations. 

While tabus represent the main distinguish feature of TS, this feature can
introduce other issues in the search process.
In particular, the use of tabus can prohibit attractive moves, or it may lead
to an overall stagnation of the search process, with a lot of moves that are
not allowed.
Hence, several methods for revoking a tabu have been
defined~\citep{glover1990tabu}, and they are commonly known as aspiration
criteria.
The simplest and most commonly used aspiration criterion consists of allowing
a move (even if it is tabu) if it results in a solution with an objective value
better than that of the current best-known solution (since the new solution has
obviously not been previously visited).

The basic TS algorithm, considering the \review{maximization} of the objective function
$f$, works as follows:
\begin{enumerate}
\item randomly select an initial solution $i$ in the search space $S$, and
set $i^* = i$ and $k=0$, where $i^*$ is the best solution so far, and $k$ the
iteration counter;
\item set $k=k+1$ and generate the subset $V$ of the \emph{admissible}
neighborhood solutions of $i$ (i.e., non-\emph{tabu} or allowed by aspiration);
\item choose the best $j$ in $V$ and set $i = j$;
\item if $f(i) > f(i^*)$, then set $i^* = i$;
\item update the \emph{tabu} and the aspiration conditions;
\item if a stopping condition is met then stop; else go to Step 2.
\end{enumerate}

The most commonly adopted conditions to end the algorithm are when the number of
iterations ($K$) is larger than a maximum number of allowed iterations, or if
no changes to the best solution have been performed in the last $N$ iterations
(as it is in our tests).

Specifically, in our experiments, we modeled for both HC and TS the possible
valid solutions in the search space as a binary adjacency matrix describing
acyclic directed graphs.
The starting point of the search is the empty graph (i.e. without any edge), and
the search is stopped when the current best solution cannot be improved with
any move in its neighborhood.

The algorithms then consider a set of possible solutions (i.e., all the directed acyclic graphs) and navigate among them by means of $2$ moves: insertion of a new edge or removal of an edge currently in the \review{structure}. We also recall that in the literature many alternatives are proposed to navigate the search space when learning the structure of Bayesian Networks, see \cite{chickering2002learning,teyssier2012ordering}. But, for the purpose of this work, we preferred to stick with the classical (and simpler) ones. 

\subsection{Genetic Algorithms}

\begin{figure*}[!t]
\begin{center}
\includegraphics[width=0.7\textwidth]{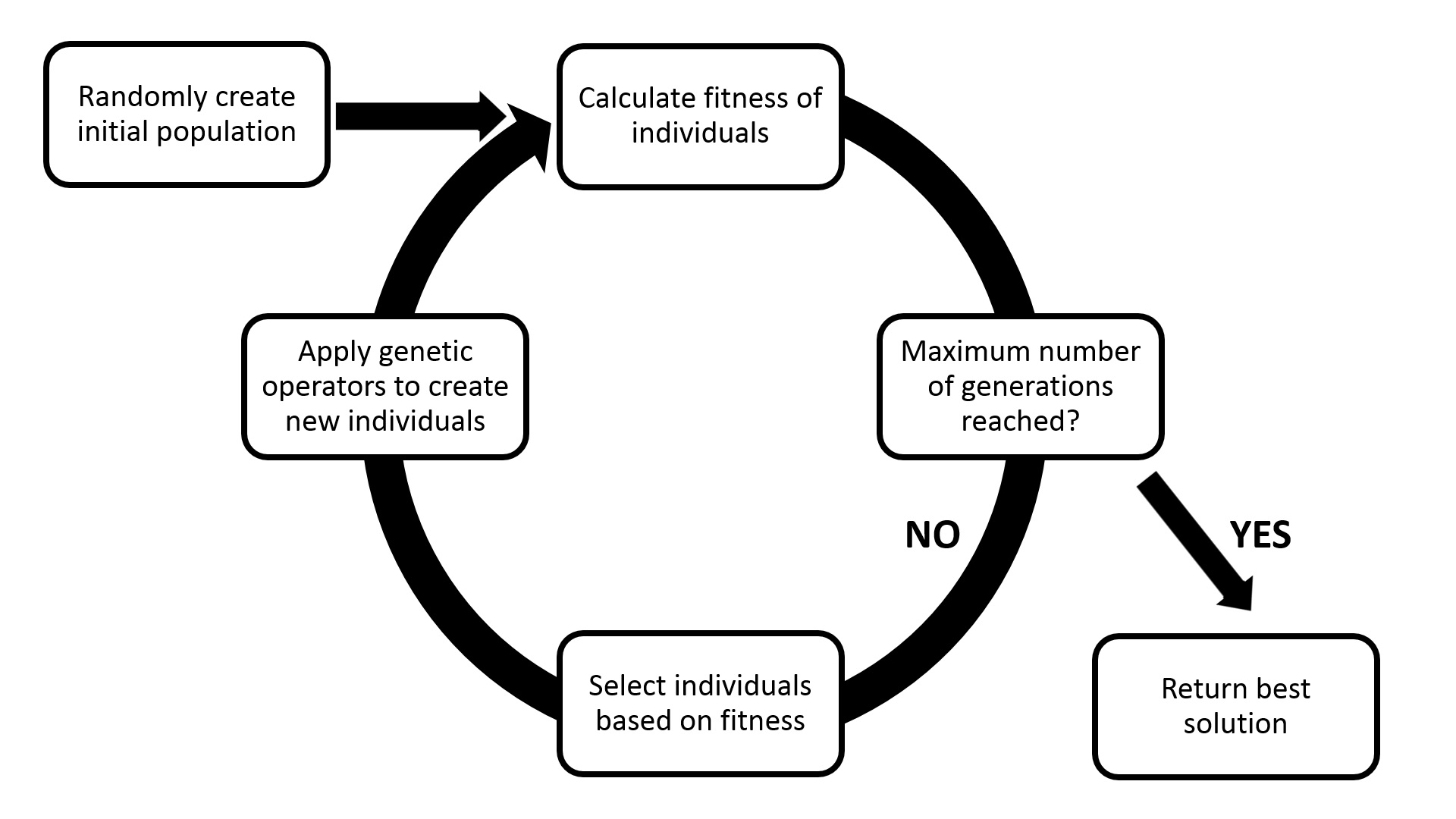} \\
\end{center}
\caption{Typical workflow of a GA algorithm.}
\label{fig:GA}
\end{figure*}

\GA (GAs)~\cite{goldberg1988genetic} are a class of computational models that
mimic the process of natural evolution.
GAs are often viewed as function optimizers although the range of problems to
which GAs have been applied is quite broad.
Their peculiarity is that potential solutions that undergo evolution are
represented as fixed length strings of characters or numbers.
Generation by generation, GAs stochastically transform sets (also called
\emph{populations}) of candidate solutions into new, hopefully improved,
populations of solutions, with the goal of finding one solution that suitably
solves the problem at hand.
The quality of each candidate solution is expressed by using a user-defined
objective function called \emph{fitness}.
The search process of GAs is shown in Figure~\ref{fig:GA}.

To transform a population of candidate solutions into a new one, GAs make
use of particular operators that transform the candidate solutions, called
genetic operators: \emph{crossover} and \emph{mutation}.
Crossover is traditionally used to combine the genetic material of two \review{candidate} solutions (parents) by swapping a part of one individual \review{(substring)} with a part of the other.
On the other hand, mutations introduce random \review{changes in the strings representing candidate solutions}. 
In order to be able to use GAs to solve a given optimization problem,
\review{candidate solutions must be encoded into strings and often}, also 
the genetic operators (crossover and mutation) \review{must be specialized for the considered context}.

GAs have been widely used to learn the BN structure considering the search
space of the DAGs.
In the large majority of the
works~\cite{cotta2002towards,van2003building,etxeberria1997analysis,kim2005robust,mascherini2005m,larranaga1996structure},
the GA encodes the connectivity matrix of the BN structure in its individuals.
This is the same approach used in our study. Our GA follows the common structure:

\begin{enumerate}
\item Generate the initial population
\item Repeat the following operations until the total number of
generations is reached:
\begin{enumerate}
\item Select a population of parents by repeatedly applying the
parent selection method
\item Create a population of offspring by applying the crossover
operator to each set of parents
\item For each offspring, apply one of the mutation operators with
a given mutation probability
\item Select the new population by applying the survivor selection
method
\end{enumerate}
\item Return the best performing individual 
\end{enumerate}

Unless stated otherwise, the following description is valid for both GA
variants (discrete and continuous).
The initialization method and the variation operators used in our GA
ensure that every individual is valid, i.e., each individual is guaranteed
to be an acyclic graph.
The initialization method creates individuals with exactly $N / 2$
connections between variables, where $N$ is the total number of variables.
A connection is only added if the graph remains acyclic.
The nodes being connected are selected with uniform probability from
the set of variables.
In the continuous variant, \review{since the input data is normalized}, the value associated with each connection is
randomly generated between $0.0$ and $1.0$, with uniform probability.
The crossover operates over two parents and returns one offspring.
The offspring is initially a copy of the first parent.
Afterward, a node with at least one connection is selected from the
second parent and all the valid connections starting from this node are
added to the offspring.

Three mutation operators are considered in our GA implementation: add
connection mutation, remove connection mutation, and perturbation mutation.
The first two are applied in both GA variants, while the perturbation
mutation is only applied in the continuous variant.
The offspring resulting from the add connection mutation operator differs
from the parent in having an additional connection.
The newly connected nodes are selected with uniform probability.
In the continuous variant, the value associated with the new connection is
randomly generated between $0.0$ and $1.0$ with uniform probability.
Similarly, the offspring resulting from the remove connection mutation
operator differs from the parent in having one less connection.
The nodes to be disconnected are selected with uniform probability.
The perturbation mutation operator applies Gaussian perturbations to the
values associated with at most $N / 2$ connections.
The total number of perturbations may be less than $N / 2$ if the
individual being mutated has less than $N / 2$ connections.
Each value is perturbed following a Gaussian distribution having mean
$0.0$ and standard deviation $1.0$.
Any resulting value below $0.0$ or above $1.0$ is bounded to $0.0$ or
$1.0$, respectively.
Regardless of the GA variant, mutation is applied with a probability of $0.25$.
When a mutation is being applied, the specific mutation operator is
selected with uniform probability from the available options (two in
the discrete variant, and three in the continuous variant).
\review{The mutation operator of the GA works by applying the perturbation to the existing values associated to each edge of the solutions. Since such modification to the value should not be highly disruptive, a common choice is to employ a Gaussian probability distribution having mean $0$ and standard deviation $1$.}

In terms of parameters, a population of size $100$ is used, with the
evolutionary process being able to perform $100$ generations.
Parents are selected by applying a tournament of size $3$.
The survivor selection is elitist in the sense that it ensures that the
best individual survives to the next generation.

\section{Results and Discussion}
\label{sec:results}

We made use of a large set of simulations on randomly generated datasets
with the aim of assessing the characteristics of the state-of-the-art
techniques for structure learning of BNs.

We generated data for both the case of discrete ($2$ values, i.e., $0$
or $1$) and continuous (we used $4$ levels, i.e., $1$, $2$, $3$, and $4$)
random variables.
\review{Notice that, for computational reasons, we discretized our continuous variables using only $4$ categories.}
For each of them, we randomly generated both the structure (i.e, $100$
weakly connected DAGs having $10$ and $15$ nodes) and the related
parameters (we generated random values in $(0,1)$ for each entry of the
conditional probability tables attached to the network structure) to build
the simulated BNs.
We also considered $3$ levels of density of the networks, namely $0.4$,
$0.6$, and $0.8$ of the complete graph.
For each of these scenarios, we randomly sampled from the BNs 
\review{several datasets of different size}, based on the number of nodes.
Specifically, for networks of $10$ variables, we generated datasets of
$10$, $50$, $100$, and $500$ samples, while for $15$ variables, we
considered datasets of $15$, $75$, $150$, and $750$ samples.
Furthermore, we also considered \review{additional noise} in the samples as a set of random
entries (both false positives and false negatives) in the dataset.
\review{We recall that, based on sample size, the probability distribution encoded in the generated datasets may different from the one subsumed by the related BN. However, here we also consider additional noise (besides the one due to sample size) due, for example, to errors in the observations. We call \emph{noise free}, the datasets in which such an additional noise is not applied.}
To this extent, we considered noise free dataset (noise rate equals $0\%$)
and dataset with an error rate of $10\%$ and $20\%$.
In total this led us to a total number of $14400$ random datasets.

For all of them, we considered both the constraint-based and the
score-based approaches to structural learning.
From the former category of methods, we considered the PC
algorithm~\cite{spirtes2000causation} and the
IAMB~\cite{tsamardinos2003algorithms}.
We recall that these methods return a partially directed graph, leaving
undirected the arcs that are not unequivocally directable.
In order to have a fair comparison with the score-based method
which returns DAGs, we randomly resolved the ambiguities, by generating
random solutions (i.e., DAGs) consistent with the statistical constraints
by PC and IAMB (that is, we select a random direction for the undirected
arcs). 

Moreover, among the score-based approaches, we consider $3$ maximum likelihood
scores, namely log-likelihood~\cite{koller2009probabilistic},
BIC~\cite{schwarz1978estimating}, and AIC~\cite{akaike1998information} (for
continuous variables we used the corresponding Gaussian likelihood scores).
For all of them, we repeated the inference on each configuration by using HC,
TS, and GA as search strategies. 

This led us to $11$ different methods to be compared, for a total of $158400$
independent experiments.
To evaluate the obtained results, we considered both false positives, FP
(i.e., arcs that are not in the generative BN but that are inferred by the
algorithm) and false negatives, FN (i.e., arcs that are in the generative BN
but that are not inferred by the algorithm).
Also, with TP (true positives) being the arcs in the generative model and TN
(true negatives) the arcs that are not in the generative model, for all the
experiments we considered the following metrics:
\begin{gather*}
Precision = \frac{TP}{TP + FP} \quad\quad\quad
Recall = \frac{TP}{TP + FN} \\\\
Specificity = \frac{TN}{TN + FP} \quad\quad
Accuracy = \frac{TP + TN}{TP + FP + TN + FN}
\end{gather*}

\subsection{Results of the Simulations}
\label{subsec:sim_results}
In this Section, we comment on the results of our simulations.
As anticipated, we computed precision, recall (sensitivity), and
specificity, as well as accuracy and \emph{Hamming} distance, to assess
the performance and the underfit/overfit trade-off of the different
approaches.
Overall, from the obtained results, it is straightforward to notice that
methods including more edges in the inferred networks are also more subject to
errors in terms of accuracy, which may also resemble a bias of this metric
that tends to penalize solutions with false positive edges rather than false
negatives.
On the other hand, since a typical goal of the problems involving
the inference of BNs is the identification of novel relations
(that is, proposing novel edges in the network), underfitting approaches could
be more effective in terms of accuracy, but less useful in practice.

With a more careful look, the first evidence we obtained from the simulations
is that the two parameters with the highest impact on the inferred networks are the density and the number of nodes (i.e.,~network size).
For this reason, we first focused our attention on these two parameters,
and we analyzed how the different tested combinations of methods and
scores behave.

\begin{figure}[!t]
\centering
\includegraphics[width=.99\textwidth]{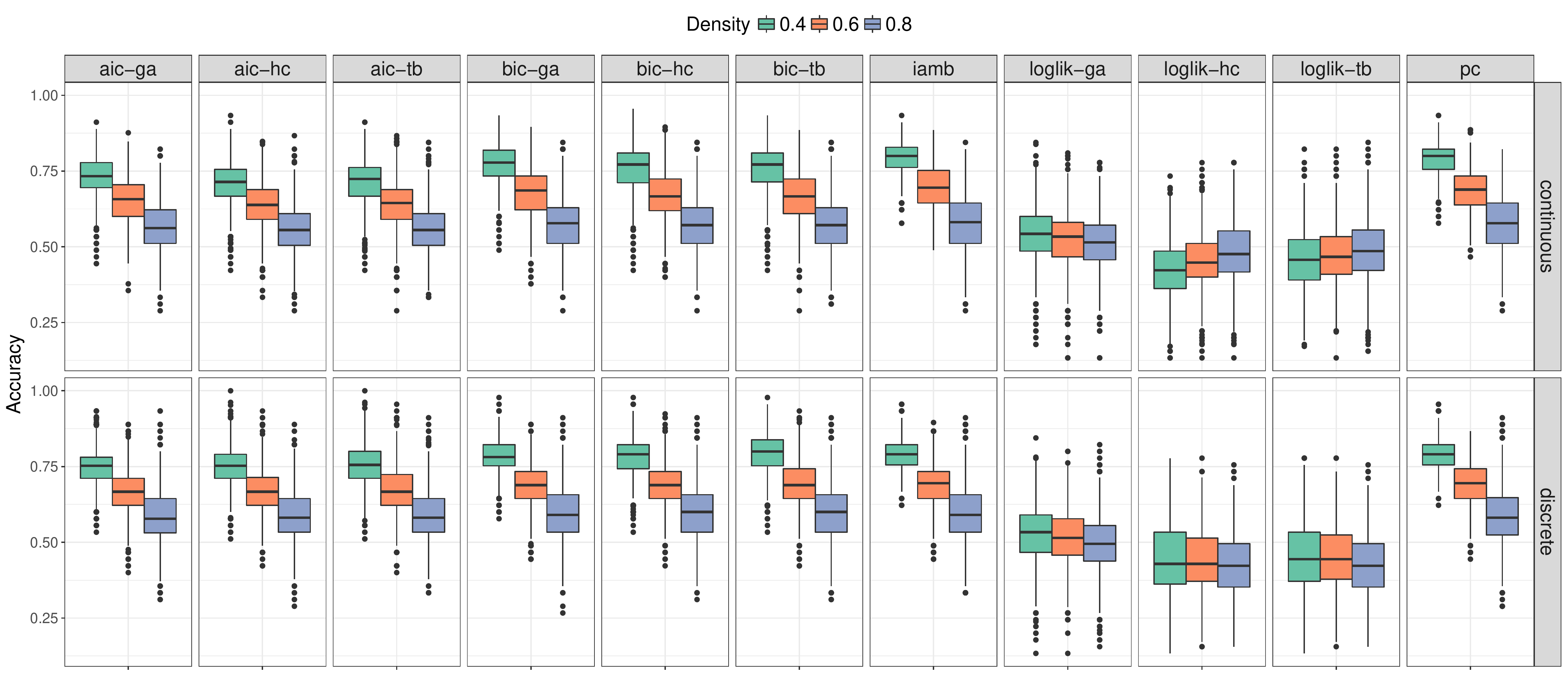}
\caption{Boxplots showing the accuracy of the results obtained by the
$11$ tests approaches on the discrete and continuous datasets. \review{In the Figure, conditional boxplot in each sub-panel reports results for structure of density $0.4$, $0.6$ and $0.8$ from left to right.}}
\label{fig:acc_dens}
\end{figure}

As shown in Figure~\ref{fig:acc_dens}, all the approaches seem to
perform better when dealing with low-density networks, in fact,
for almost all the methods the accuracy is higher for density equal
to $0.4$, while it is lower for density equal to $0.8$.
Since each edge of the network is a parameter that has to be ``learned'',
it is reasonable to think that, the more edges are present in the BN, the harder becomes the problem to be solved when learning it.
Moreover, we can also observe that the results in terms of accuracy obtained for BN of discrete variables are slightly
better than those achieved on the continuous ones.
To this extent, the only outlier is the loglik score combined with
\HC and with \TS on datasets with continuous variables, for which the
trend is the opposite, w.r.t.~that of all the other approaches.
In fact, in both these cases, the accuracy is higher on high-density datasets, and this is likely due to a very high overfit of the approach.
It is interesting to notice that \GA (combined with the loglik score) is
less affected by this problem, and this trend will also be shown in the
next analyses.

\begin{figure}[!t]
\centering
\includegraphics[width=.99\textwidth]{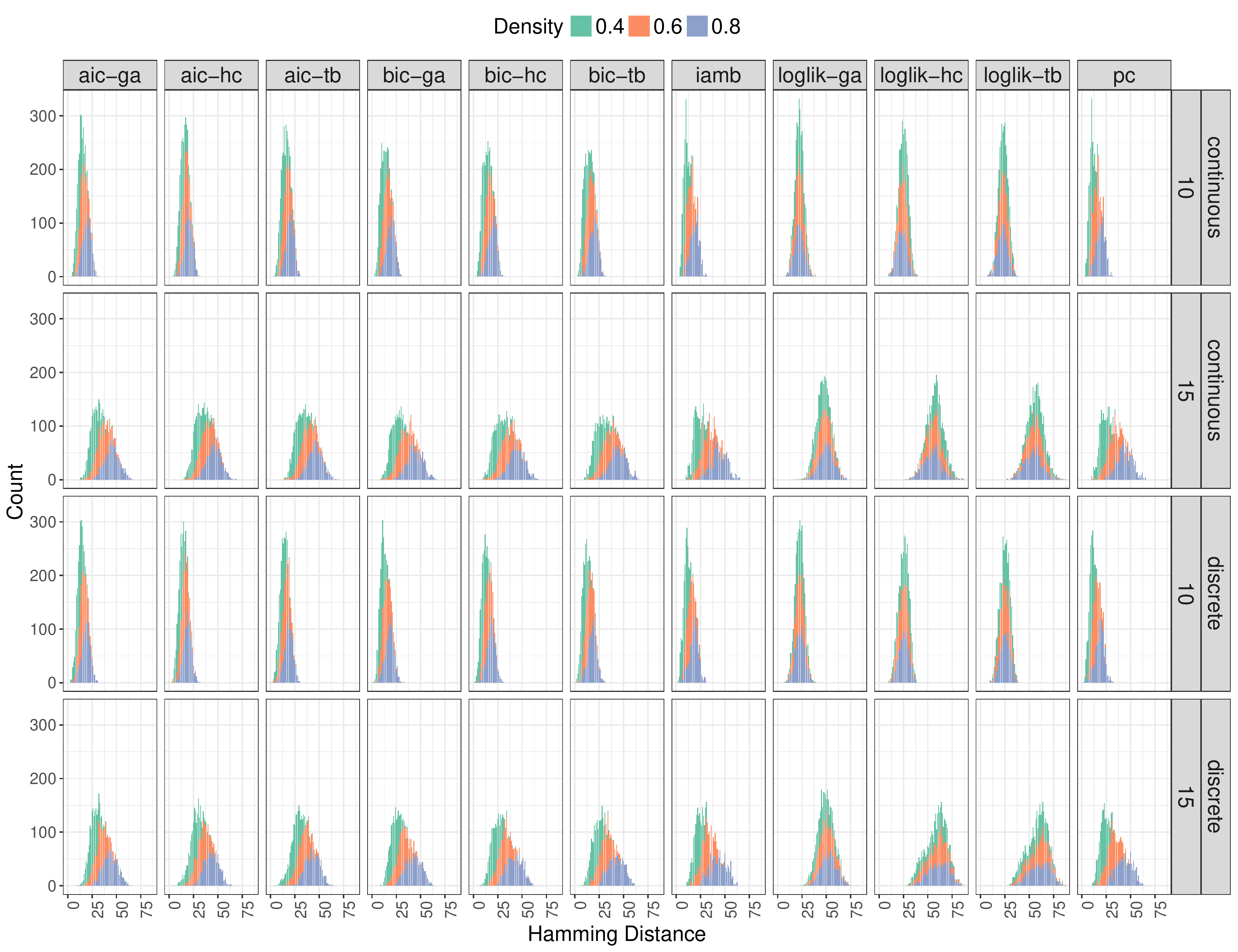}
\caption{Bar plots showing the \emph{Hamming} distances of the
solutions obtained by the $11$ tested approaches, grouped by type
of dataset (discrete and continuous) and number of nodes of the
network ($10$ and $15$), while colors represent densities ($0.4$,
$0.6$, and $0.8$).}
\label{fig:hamming}
\end{figure}

In addition to the accuracy, we also computed
the \emph{Hamming} distance between the reconstructed solutions and the
BNs used to generate the data, in order to quantify the errors of the
inference process.
This analysis showed that, besides the network density, also
the number of nodes (parameters to be learned) influences the results,
as reported in Figure~\ref{fig:hamming}.
It is interesting to observe that, in the adopted experimental setup,
we set the number of samples to be proportional to the number of nodes
of the network, that is, for $10$ nodes, in the same configurations we
have a lower number of samples, compared to the ones for $15$ nodes.
From a statistical point of view alone, we would expect the problem to be
easier when having more samples to build the BN, since this
would lead to more statistical power, and, intuitively, should compensate
for the fact that with $15$ nodes we have more parameters to learn than
the ones for the case of $10$ nodes.
While this may be the case (in fact we have similar accuracy for both
$10$ and $15$ nodes), we also observe a constantly higher \emph{Hamming}
distance with more nodes.
In fact, when dealing with more variables we observed a shift in the
performance, that is, even when the density is low, we observe more errors manifesting with higher values in terms of \emph{Hamming} distance.
This is due to the fact that, when increasing the number of variables,
we also increase the complexity of the solutions. 

\begin{figure}[!t]
\centering
\includegraphics[width=.99\textwidth]{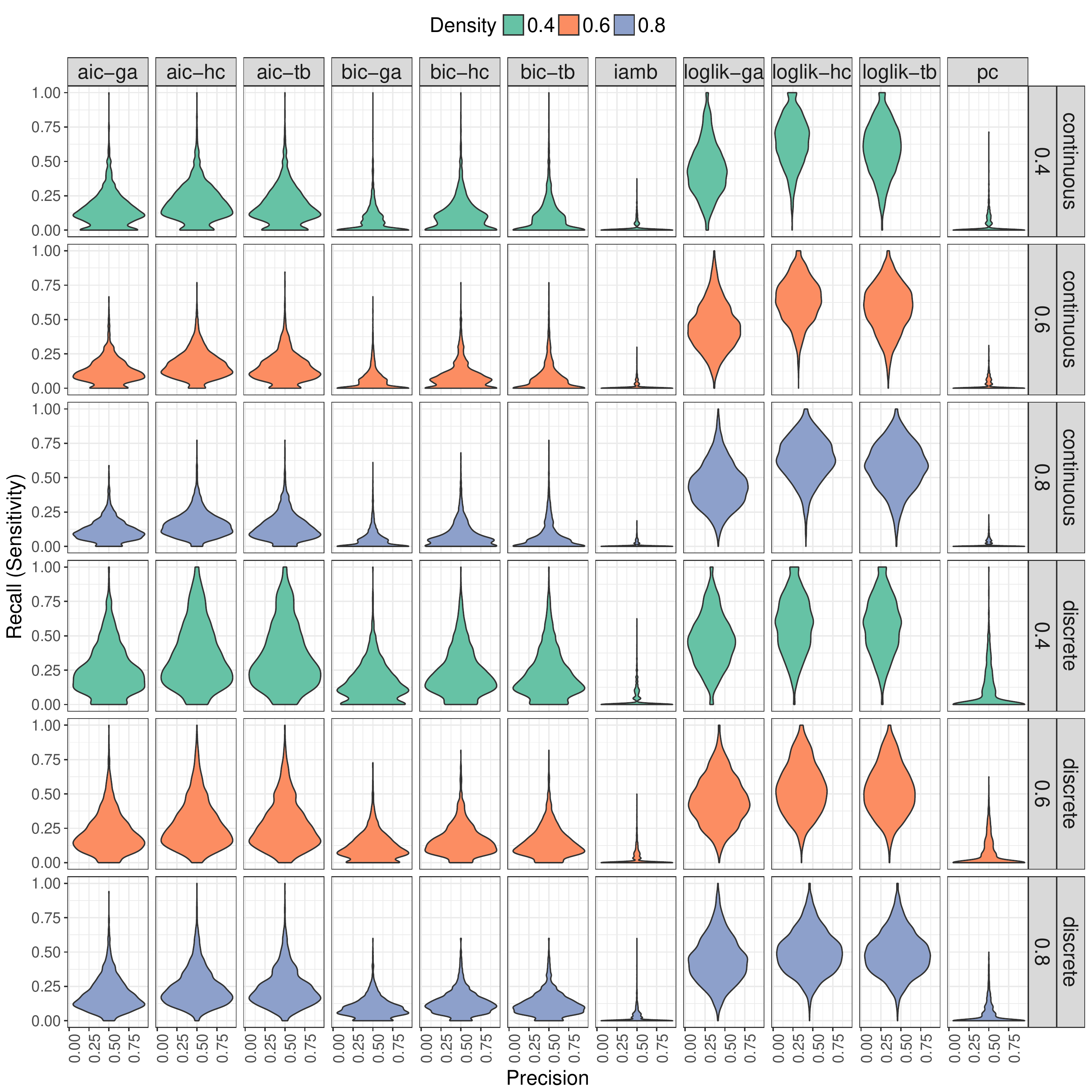}
\caption{Violin plots showing Precision and Recall of the
solutions obtained by the $11$ tested approaches, grouped by the type
of dataset (discrete and continuous) and density values ($0.4$,
$0.6$, and $0.8$).}
\label{fig:precrec_violin}
\end{figure}

\begin{figure}[!t]
\centering
\includegraphics[width=.99\textwidth]{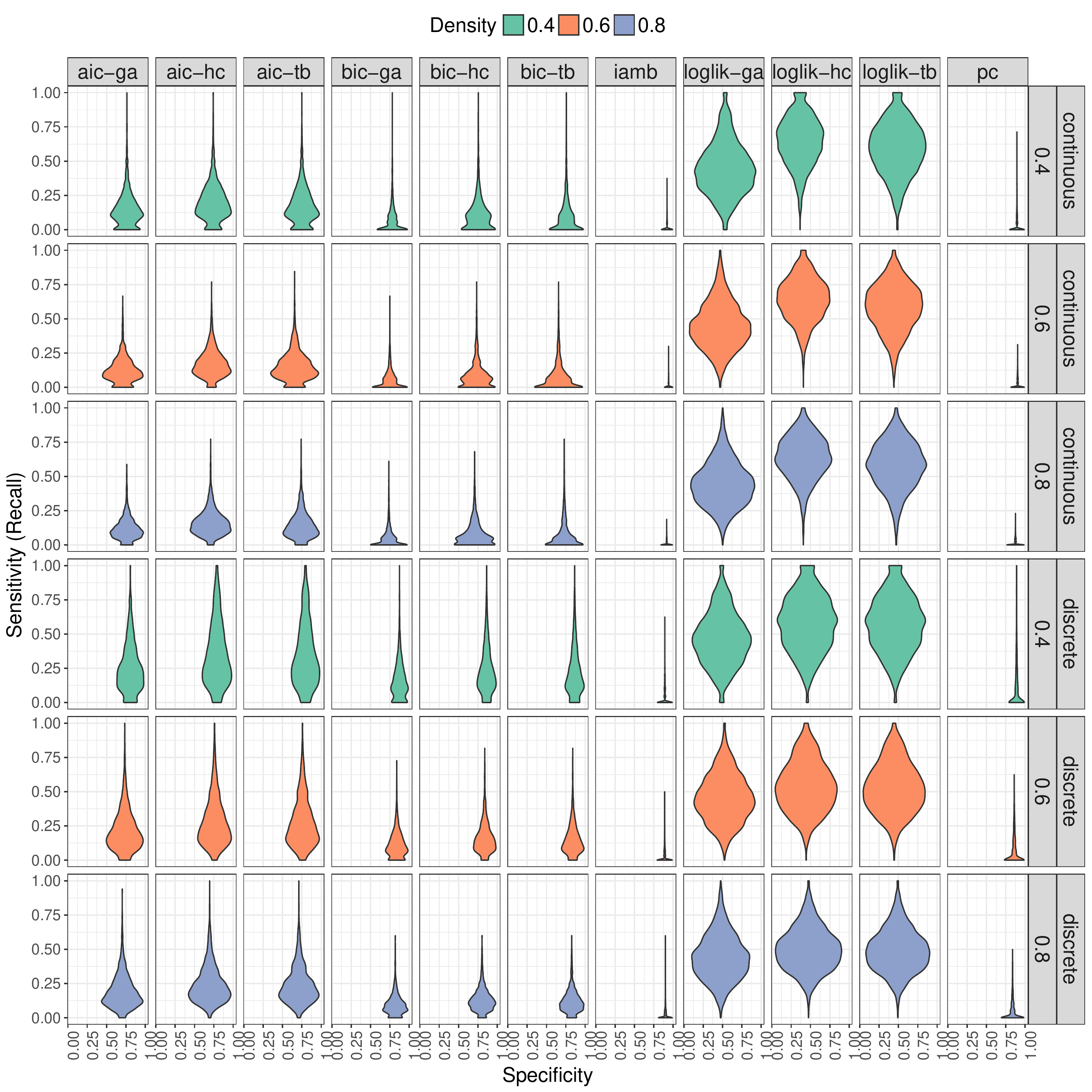}
\caption{Violin plots showing Specificity and Sensitivity of the
solutions obtained by the $11$ tested approaches, grouped by the type
of dataset (discrete and continuous) and density values ($0.4$,
$0.6$, and $0.8$).}
\label{fig:sensspec_violin}
\end{figure}

A further analysis we performed was devoted to assessing the impact of
both overfit and underfit.
As it is possible to observe from the plots in
Figure~\ref{fig:precrec_violin} and Figure~\ref{fig:sensspec_violin},
we obtain two opposite behaviors, namely, the combinations of scores and
search strategies show evident different trends in terms of \emph{sensitivity}
(\emph{recall}), \emph{specificity}, and \emph{precision}.
In details, \emph{iamb} and \emph{pc} tend to underfit, since they both
produce networks with consistently low density.
While they achieve similar (high) results in terms of accuracy (see
Figure~\ref{fig:acc_dens}), their trend toward underfit does not make
them suited for the identification of novel edges, rather being more
indicated for descriptive purposes.
On the other hand, the \emph{loglik} score, (mostly) independently from
the adopted search technique, consistently overfits.
Both these behaviors can be observed in the violin plots of 
Figure~\ref{fig:precrec_violin} and Figure~\ref{fig:sensspec_violin}:
for each density, we can observe that iamb/pc results have very low
\emph{recall} values, while the distribution of the results of the
loglik score is centered on higher values.

The other two scores, i.e., AIC and BIC, present a better trade-off
between overfit and underfit, with BIC being less affected by the
overfit, while AIC reconstructing slightly denser networks.
Once again, this trade-off between the two regularizators is well-known
in the literature and points to AIC being more suited for predictions while
BIC for descriptive purposes.

Another relevant result of our analysis is the characterization of the performance of \GA
in terms of \emph{sensitivity} (\emph{recall}).
Specifically, it must be noticed that for all the three regularizators,
\GA achieve similar results of both \emph{precision} and
\emph{specificity} if compared with both \HC and \TS, but in terms
of \emph{sensitivity} (\emph{recall}) it presents reduced overfit.
In fact, as it is possible to observe in the plots of
Figure~\ref{fig:precrec_violin} and Figure~\ref{fig:sensspec_violin}
for each score (i.e.,~loglik, AIC, and BIC), the \emph{sensitivity}
results of \GA are lower than that of both \HC and \TS, highlighting
the reduced impact of overfit.

In summary, we can draw the following conclusions.

\begin{itemize}
\item $(i)$ dense networks are harder to learn, see Figure~\ref{fig:acc_dens};
\item $(ii)$ the number of nodes effects the complexity of the solutions,
leading to higher \emph{Hamming} distance (number of errors) even if with
similar accuracy, see Figure~\ref{fig:acc_dens} and Figure~\ref{fig:hamming};
\item $(iii)$ networks with continuous variables are harder to learn
compared to the binary ones, see Figure~\ref{fig:acc_dens};
\item $(iv)$ \emph{iamb} and \emph{pc} algorithms tend to underfit, while
\emph{loglik} overfits, see Figure~\ref{fig:precrec_violin} and
Figure~\ref{fig:sensspec_violin};
\item $(v)$ \GA tend to reduce underfit, see Figure~\ref{fig:precrec_violin} and Figure~\ref{fig:sensspec_violin}. 
\end{itemize}

We conclude the section by providing $p-$values by Mann–Whitney $U$
test~\cite{mann1947test} in support of such claims in
Table~\ref{tab:pvalue_results}.
The claim for greater and less values, i.e., accuracy, are performed with
the one-tail test alternatives.
The tests are performed on all the configurations of our simulations. 

\begin{table}[!t]
\centering
\begin{tabular}{| c | c | c | c | c | c |}
\hline
\textbf{Comparison} & \textbf{Metric} & \textbf{Test} & \textbf{$p-$value} & \textbf{Mean $1$} & \textbf{Mean $2$} \\ \hline
density $0.4$/$0.8$ & accuracy & greater & $< 2.2e-16$ & $0.68$ & $0.55$ \\ \hline
nodes $10$/$15$ & accuracy & less & $< 2.2e-16$ & $0.61$ & $0.62$ \\ \hline
nodes $10$/$15$ & Hamming & less & $< 2.2e-16$ & $17.40$ & $39.54$ \\ \hline
discr./contin. & accuracy & greater & $< 2.2e-16$ & $0.62$ & $0.61$ \\ \hline
iamb/hc loglik & sensitivity & less & $< 2.2e-16$ & $0.02$ & $0.59$ \\ \hline
iamb/hc loglik & specificity & greater & $< 2.2e-16$ & $0.98$ & $0.38$ \\ \hline
ga/hc & sensitivity & less & $< 2.2e-16$ & $0.24$ & $0.32$ \\ \hline
ga/hc & specificity & greater & $< 2.2e-16$ & $0.79$ & $0.71$ \\
\hline
\end{tabular}
\label{tab:pvalue_results}
\caption{Summary of the major findings.
The results of the Mann–Whitney $U$ test \cite{mann1947test} (one-tail
test alternatives) in support of the results of our simulations are shown.
The tests are performed on all the settings, with the comparisons as
described in the first column of the table.
We also show the considered metric, the adopted alternative of the one-tail
test, the obtained $p-$value, and the mean of the two compared distributions
of results.}
\end{table}

\section{Conclusions}
\label{sec:conclusions}

Bayesian Networks are a widespread technique to model dependencies among
random variables.
A challenging problem when dealing with BNs is the one of learning their
structures, i.e., the statistical dependencies in the data, which may
sometime pose a serious limit to the reliability of the results. 

Despite their extensive use in a vast set of fields, to the best of our
knowledge, a quantitative assessment of the performance of different
state-of-the-art methods to learn the structure of BNs has never been
performed.
In this work, we aim at going in the direction of filling this gap and
we presented a study of different state-of-the-art approaches for
structural learning of Bayesian Networks on simulated data, considering
both discrete and continuous variables.

To this extent, we investigated the characteristics of $3$ different
likelihood score, combined with $3$ commonly used search strategies schemes
(namely, \GA, \HC, and \TS), as well as $2$ constraint-based techniques
for the BN inference (i.e., iamb and pc algorithms).

Our analysis identified the factors having the highest impact on the performance,
that is, density, number of variables, and variable type (discrete vs
continuous).
In fact, as shown here, these settings affect the number of parameters to be
learned, hence complicating the optimization task.
Furthermore, we also discussed the \review{overfit}/underfit trade-off of the different
tested techniques with the constraint-based approaches showing trends toward underfitting, and the loglik score showing high overfit.
Interestingly, in all the configurations, \GA showed evidence of reducing
the overfit, leading to denser structures. 

Overall, we place our work as a starting effort to better characterize the
task of learning the structure of Bayesian Networks from data, which may
lead in the future to a more effective application of this approach.
\review{In particular, we focused on the more general task of learning the structure of a BN~\cite{koller2009probabilistic}, and we did not dwell on several interesting domain-specific topics, which we leave for future investigations~\cite{drton2017structure,stefanini2014chain,heinze2018causal}.}

\section*{Acknowledgments}
This work was also financed through the Regional Operational Programme CENTRO2020 within the scope of the project CENTRO-01-0145-FEDER-000006.



\section*{References}
\bibliographystyle{model1-num-names}
\bibliography{biblio.bib}







\end{document}